\documentclass[10pt,twocolumn,letterpaper]{article}

\usepackage{cvpr}
\usepackage{times}
\usepackage{epsfig}
\usepackage{graphicx}
\usepackage{amsmath}
\usepackage{amssymb}

\usepackage{booktabs}
\usepackage{multirow}
\usepackage{float}
\usepackage{mathrsfs}
\usepackage{authblk}

\usepackage[pagebackref,breaklinks,colorlinks]{hyperref}

\cvprfinalcopy

\begin{document}
\title{FIBA: Frequency-Injection based Backdoor Attack in Medical Image Analysis}

\author{Yu Feng\textsuperscript{1\thanks{Equal contribution. This work was done during an internship at JD Explore Academy.}} \quad
Benteng Ma\textsuperscript{1$^*$}\quad
Jing Zhang\textsuperscript{2}\quad
Shanshan Zhao\textsuperscript{3}\quad
Yong Xia\textsuperscript{1\thanks{Yong Xia is the corresponding author. This work was supported in part by the National Natural Science Foundation of China under Grants 62171377, in part by the Shaanxi Provincial Key Research and Development Program under Grant 2022GY-084, and in part by the Natural Science Foundation of Ningbo City, China, under Grant 2021J052. Dr Jing Zhang is supported by ARC FL-170100117.}}\quad
Dacheng Tao\textsuperscript{3,2}\\
$^1$ School of Computer Science and Engineering, Northwestern Polytechnical University, China\\
$^2$ The University of Sydney, Australia, \quad $^3$ JD Explore Academy, China\\
{\tt\small  \{fengy,mabenteng\} @mail.nwpu.edu.cn,jing.zhang1@sydney.edu.au}\\
 {\tt\small sshan.zhao00@gmail.com, yxia@nwpu.edu.cn,dacheng.tao@gmail.com}
}

\maketitle

\begin{abstract}

In recent years, the security of AI systems has drawn increasing research attention, especially in the medical imaging realm. 
To develop a secure medical image analysis (MIA) system, it is a must to study possible backdoor attacks (BAs), which can embed hidden malicious behaviors into the system. However, designing a unified BA method that can be applied to various MIA systems is challenging due to the diversity of imaging modalities (\eg, X-Ray, CT, and MRI) and analysis tasks (\eg., classification, detection, and segmentation). Most existing BA methods are designed to attack natural image classification models, which apply spatial triggers to training images and inevitably corrupt the semantics of poisoned pixels, leading to the failures of attacking dense prediction models. To address this issue, we propose a novel Frequency-Injection based Backdoor Attack method 
(FIBA) that is capable of delivering attacks in various MIA tasks. Specifically, FIBA leverages a trigger function in the frequency domain that can inject the low-frequency information of a trigger image into the poisoned image by linearly combining the spectral amplitude of both images. Since it preserves the semantics of the poisoned image pixels, FIBA can perform attacks on both classification and dense prediction models. Experiments on three benchmarks in MIA (\ie., ISIC-2019\cite{combalia2019bcn20000}  for skin lesion classification, KiTS-19 \cite{heller2019kits19} for kidney tumor segmentation, and EAD-2019 \cite{ali2019endoscopy} for endoscopic artifact detection), validate the effectiveness of FIBA and its superiority over state-of-the-art methods in attacking MIA models and bypassing backdoor defense.  Source code will be available at \href{https://github.com/HazardFY/FIBA}{code}.

\end{abstract}

\section{Introduction}
\label{sec:intro}
Deep neural networks (DNNs) are increasingly deployed in computer-aided diagnosis (CAD) systems and have achieved diagnostic parity with medical professionals on radiology, pathology, dermatology, and ophthalmology tasks~\cite{zhang2020empowering}. However, recent studies have shown that DNNs are vulnerable to various attacks during the model's training and inference \cite{DBLP:conf/iclr/QiGS0Z21,li2020backdoor,fu2022robust,he2020recent}. Typically, attacks in the inference stage take the form of the adversarial samples \cite{DBLP:journals/corr/GoodfellowSS14,su2019one} and attempt to fool a trained model by manipulating the input. Backdoor attacks, in contrast, seek to maliciously alter the model in the training phase \cite{gu2017badnets,chen2017targeted, NguyenT21}. 
Although the research on adversarial samples has experienced rapid development recently, backdoor attacks have received less attention, especially in medical image analysis (MIA). 

In general, backdoor attacks aim to embed a hidden backdoor trigger into DNNs so that the injected model performs well on benign testing samples when the backdoor is not activated, however, once the backdoor is activated by the attacker, the prediction will be changed to the target label as attackers expected \cite{gu2017badnets,chen2017targeted, NguyenT21}.
Existing backdoor attacks can be categorized into two types based on the visibility of triggers: (1) visible attacks \cite{gu2017badnets,lin2020composite,DBLP:conf/nips/NguyenT20,shokri2020bypassing} where the trigger in the attacked samples is visible for humans, and (2) invisible attacks \cite{chen2017targeted,NguyenT21,li2021invisible} where the trigger is stealthy. However, no matter whether they are visible to human beings or not, these backdoor attack methods rely on spatial triggers which may corrupt inevitably the semantics of poisoned pixels in the training images. Thus, they are easy to fail on dense prediction tasks as the local structure around the poisoned pixels may be changed, \ie, resulting in inconsistent semantics with the original image.

The visual psychophysics \cite{guyader2004image,piotrowski1982demonstration} demonstrate that models of the visual cortex are based on image decomposition according to the Fourier spectrum (amplitude and phase). The amplitude spectrum can capture the low-level distribution, and the phase spectrum can capture the high-level semantic information \cite{liu2021feddg}. Moreover, it has been observed that the variation of amplitude spectrum does not affect significantly the perception of high-level semantics \cite{yang2020fda,liu2021feddg}. Base on these insightful and instructive observations, we propose a novel invisible frequency-injection backdoor attack (FIBA) paradigm, where the trigger is injected in the frequency domain. Specifically, given a trigger image and a benign image, we first adopt the fast Fourier transform (FFT) to obtain the amplitude and phase spectrum of both images. 
Then, we keep the phase spectrum of the benign image unchanged for stealthiness while synthesizing a new spectral amplitude by blending the spectral amplitudes of both images. Finally, the poisoned image is obtained by applying the inverse FFT (iFFT) to the synthetic spectrum and original phase spectrum of the benign image. Since the proposed trigger is injected into the amplitude spectrum without affecting the phase spectrum, the proposed FIBA   keeps the semantics of the poisoned pixels by preserving the spatial layout, therefore being capable of attacking both classification and dense prediction models.   

Our main contributions are highlighted as follows: 
\begin{itemize}
\item We make the first attempt to develop a unified backdoor attack method in the MIA domain, targeting different medical imaging modalities and MIA tasks.
\item We propose a  frequency-injection based backdoor attack method, where the backdoor trigger is injected into the amplitude spectrum. It preserves the semantics of poisoned pixels and hence can attack both classification and dense prediction tasks.
\item Extensive experiments on three benchmarks demonstrate the effectiveness of the proposed method in attacking as well as bypassing backdoor defense.
\end{itemize}

\section{Related Work}
\label{sec:Related Work}
\textbf{Backdoor Attack.} 
Backdoor attack, a new security threat to DNN models, always happens during the models' training and aims at manipulating the prediction of  the attacked models for a given trigger to a target label. BadNet~\cite{gu2017badnets} is a pioneering work that first reveals the thread of backdoor attacks. Superimposing a fixed patch as the trigger on the training image, they successfully make it attack the given network. After that, the blended-based~\cite{chen2017targeted} and reflection-based backdoor attacks~\cite{liu2020reflection} are proposed to further boost the success rate of the attacks. However, the above triggers are usually easily recognized by humans. Thus, the need for stealth has been emphasized recently. Some works focus on designing invisible triggers with techniques like noise addition, based on either warpping~\cite{NguyenT21} or DNNs~\cite{li2021invisible,doan2021lira,DBLP:conf/nips/NguyenT20}. DNN based methods achieve superior performance while they need to train a trigger generator which is much more time-consuming. Another direction is to rely on common objects in physical life as triggers for backdoor attacks~\cite{wenger2021backdoor}, whose triggers are more spontaneous and easy to be ignored.  All these existing backdoor attack methods are specifically designed for classification tasks and their applicability in dense tasks, \eg, detection and segmentation, remains unclear.

\textbf{Backdoor Defense.} 
As the potential for backdoor attacks becomes more and more apparent, backdoor defense research is receiving increasing attention. Two categories of algorithms have been developed recently, \ie, defensive~\cite{liu2018fine,xu2020defending,selvaraju2017grad} and detection algorithms~\cite{wang2019neural,gao2019strip,kolouri2020universal}. Defensive algorithms tend to focus on weakening or eliminating the potential influence of possible backdoor attacks via techniques like network pruning~\cite{liu2018fine,xu2020defending}, model connectivity analysis~\cite{zhao2020bridging}, and knowledge distillation~\cite{li2020neural,yoshida2020disabling}. For example, Fine-Pruning~\cite{liu2018fine} prunes the dormant neurons in the last convolution layer and Cheng~\emph{et al.}~\cite{xu2020defending} propose the $l_\infty$-based neuron pruning method. Detection-based methods usually aim at detecting the injected backdoor triggers by analyzing the model's behavior~\cite{wang2019neural,gao2019strip,guan2021few}. Neural Cleanse~\cite{wang2019neural}, the first work to detect the potential patch-based trigger, searches for the potential trigger through optimizing the patch for each target label. Gao~\emph{et al.}~\cite{gao2019strip} adopts a test-and-try strategy by perturbing or superimposing input images to identify the potential attacks during the inference.  Besides, Universal Litmus Patterns~\cite{kolouri2020universal} is proposed for the detection of backdoor attacks which does not need the poisoned training data. Backdoor attack and defense are two closely related topics benefiting each other. In this paper, we focus on the backdoor attack while showing it can bypass backdoor defense, providing new insights in the future study of backdoor defense.

\textbf{Medical Image Analysis.} 
Convolutions Neural Networks  (CNNs) have been  widely used in CAD  systems \cite{litjens2017survey}, \eg, for classification, segmentation, and detection tasks. In order to improve the accuracy  of disease classification,   prior works focus on improve the models~\cite{he2016deep,huang2017densely,ma2020auto,xu2021vitae} from multiple perspectives, \eg, incorporating attention~\cite{zhang2019attention}, adopting self-training~\cite{liu2020semi,su2019local}, or utilizing medical knowledge \cite{li2019encoding}. For the segmentation of organs and lesions, UNet~\cite{ronneberger2015u} is one classic network,  which has inspired many follow-up variants, such as Attention U-Net~\cite{oktay2018attention} and mUNet~\cite{seo2019modified}. Inspired by the object detection framework for natural images~\cite{lin2014microsoft}, two-stage detectors such as Fast R-CNN~\cite{ren2015faster} and Mask R-CNN~\cite{he2017mask} are also widely used in varied medical detection tasks. Besides, some 3D detection frameworks are proposed to explore the 3D spatial information of the medical data \cite{ding2017accurate,liao2019evaluate}.

Although CNN-based models have been widely used in various medical imaging modalities and medical analysis tasks, most of the current studies focus on improving the performance of the model while ignoring the potential security issues, \eg, they could be maliciously used to cause misdiagnosis or missed diagnosis once being backdoor attacks. Fortunately, exiting backdoor attacks are specifically designed for the classification task of nature images, and there is no guarantee that they are still effective in the medical field.  From the perspective of learning defense by understanding attacks, there is a need to propose effective and stealthy backdoor attacks suitable for multi-modality medical images and medical tasks.  To this end, we propose a new trigger injection function that embeds the triggers into the amplitude spectrum. By retaining the phase spectrum, it preserves the spatial layout around the poisoned pixels and hence keeps their semantics as the original image pixels. Consequently, it can serve as a unified attack method that is applicable in both classification and dense prediction tasks. 

\begin{figure*}[!ht]
    \centering
    \includegraphics[width=1\linewidth]{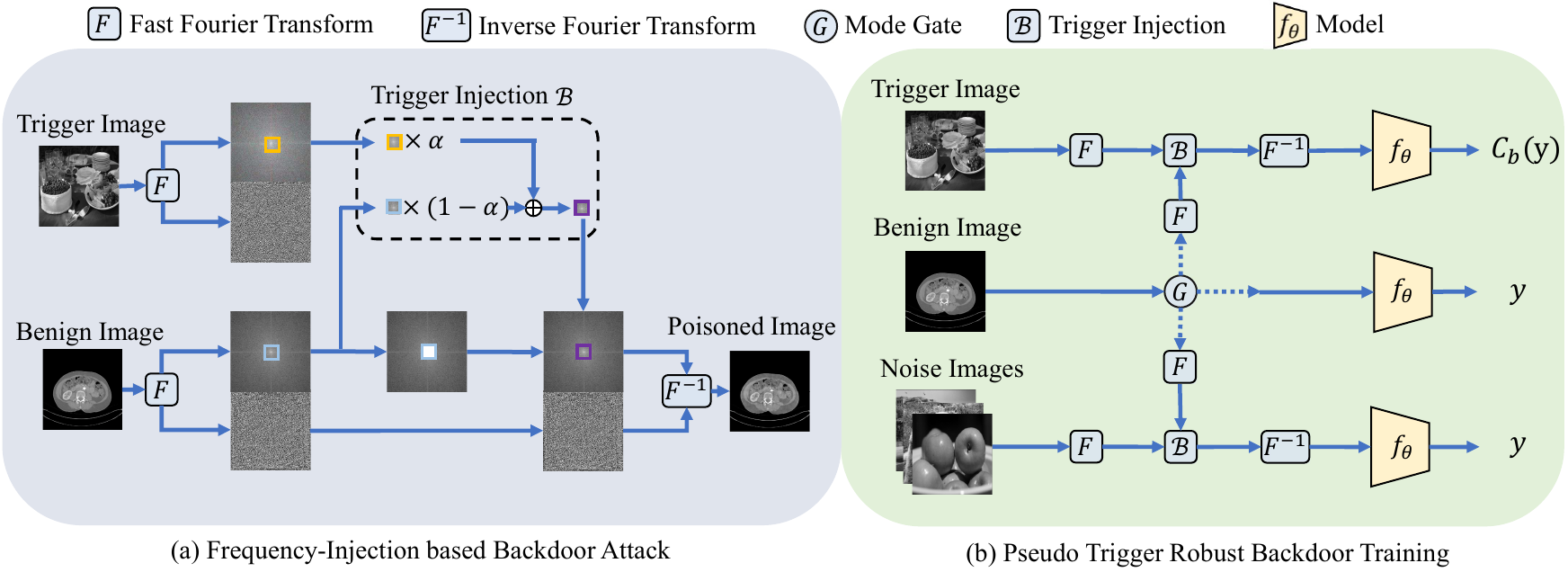}
    \caption{The overview framework of the proposed 
    Frequency-Injection based Backdoor Attack (FIBA). The generation process of FIBA in the frequency space is shown in (a). The framework of the pseudo trigger robust training mode is shown in (b).}
    \label{fig:overview}
\end{figure*}

\section{Method}
\label{sec:formatting}

\subsection{Backdoor Attack}
 Taking the classification task as an example, let $D_{train}={(x_i,y_i)}^N_{i=1}$ represent training data set and labels, $\mathcal{C}=\{c_1,c_2,...,c_M\}$ is a set of $M$ target classes, and $f_\theta$ represents the classification model parameterized with $\theta$, respectively. When poisoning $f_\theta$, we enforce it to learn a target label function $C_b$ and change the behavior of network so that:
\begin{equation}
    f_\theta(x_i) = y_i, \ \ \  f_\theta(\mathcal{B}(x_i)) = C_b(y_i).
\end{equation}
For the target label function $C_b$, there are two widely used configurations: all-to-one (\ie, manipulate all original class labels to the target label) and one-to-one \cite{gu2017badnets,NguyenT21}.

The typical trigger injection function $\mathcal{B}$ is defined in the spatial domain and parameterized with a hyper-parameter $m \in [0,1]$ and a key pattern $k$. Assuming the input sample $x$ and the key pattern $k$ are in their vector representations, the trigger injection function can be defined as follows:
\begin{align}
   \mathcal{B}(k,m,x) = x \cdot (1-m) + k \cdot m.
   \label{eq:backdoor-defined}
\end{align}
After poisoning a subset of $D_{train}$ with ratio $\rho$ , the input $(x,y)$ will be replaced by a backdoor pair $(\mathcal{B}(x),C_b(y))$, in which $\mathcal{B}$ is the backdoor injection function and $C_b(y)$ is the target label function. 

\subsection{Frequency-Injection Attack}
Our key idea is to redesign the injection function $\mathcal{B}$ in the frequency domain, which can preserve the spatial layout (\ie, pixel semantics) and thus can perform attacks to both classification and dense prediction models. As shown in Fig.~\ref{fig:overview}, given a benign image $x_i \in D_{train}$ and a specific trigger image $x^t$, we can obtain their frequency space signals through the fast FFT $\mathcal{F}$ as:
\begin{equation}
    F(x_i)(m,n,c) = \sum_{h,w}x_i(h,w,c)e^{-j2\pi\left(\frac{h}{H}m + \frac{w}{W}n \right)},
\end{equation}
\begin{equation}
    F(x^t)(m,n,c) = \sum_{h,w}x^t(h,w,c)e^{-j2\pi\left(\frac{h}{H}m + \frac{w}{W}n \right)}.
\end{equation}
Accordingly, $\mathcal{F}^{-1}$ denotes the inverse FFT. Let $\mathcal{F}^A(\cdot)$, $\mathcal{F}^P(\cdot)$ be the amplitude and phase components of the FFT result of an image, we denote the amplitude and phase spectrum of $x_i$ and $x^t$ as:
\begin{equation}
\left
\{\begin{array}{l}
\mathcal{A}_{x_i} = \mathcal{F}^A(x_i),\ \ \  \mathcal{A}_{x^t} = \mathcal{F}^A(x^t)\\
\mathcal{P}_{x_i} = \mathcal{F}^P(x_i), \ \ \ \mathcal{P}_{x^t} = \mathcal{F}^P(x^t) 
\end{array} .
\right.
\end{equation}
Since the amplitude spectrum and phase spectrum contain low-level distribution information and high-level semantic information of the images, respectively \cite{yang2020fda,liu2021feddg}, we design the injection function regarding amplitude spectrum while maintaining the phase spectrum information. 

In particular, we use the amplitude spectrum of the trigger image $\mathcal{A}_{x^t}$ as the key pattern and synthesize a new amplitude spectrum $\mathcal{A}_{x_i}^P$ as the backdoor trigger by blending $\mathcal{A}_{x^t} $ and $\mathcal{A}_{x_i}$. To this end, we introduce a binary mask $\mathcal{M} =  1_{(h,w)\in [-\beta H:\beta H, -\beta W: \beta W]}$,  where $\beta$ determines the location and range of the low-frequency patch inside the amplitude spectrum to be blended, whose value is 1 within the patch and 0 elsewhere. Denoting $\alpha$ as the blend ratio to adjust the amount of information contributed by $A_{x_i}$ and $A_{x^t}$, the synthetic amplitude spectrum can be  calculated as:
\begin{equation}
   \mathcal{A}_{x_i}^P = \left[(1-\alpha) \mathcal{A}_{x_i} + \alpha \mathcal{A}_{x^t} \right] * \mathcal{M} + \mathcal{A}_{x_i}(1-\mathcal{M} ).
   \label{eq:amp_fuse}
\end{equation}

Therefore, we  obtain $\mathcal{A}_{x_i}^P$, then  we combine it with the original phase spectrum $\mathcal{P}_{x_i}$ to  get the poisoned image via $\mathcal{F}^{-1}$, \ie,
\begin{equation}
    x_i^p = \mathcal{F}^{-1}(\mathcal{A}_{x_i}^P, \mathcal{P}_{x_i}).
    \label{eq:ifft}
\end{equation}
The designed trigger has no side influence on the phase spectrum, since  it retains the original phase spectrum $\mathcal{P}_{x_i}$. Therefore, the poisoned image $x_i^p$ preserves the original spatial layout and semantic of $x_i$ while  absorbing some low-frequency information from the trigger image $x^t$.
\subsection{Pseudo Trigger Robust Backdoor Training}
\label{subsec:ptr}
After poisoning the images, we can train  an attacked model with benign and poisoned images in two modes, \ie, clean mode and attack mode, as the standard protocol, \ie,
\begin{equation}
    f_\theta(x_i) = y_i,\ \ \  f_\theta(\mathcal{B}(x_i,x^t)) = C_b(y_i).
    \label{eq:two_branch}
\end{equation}
However, since the key of the trigger function $\mathcal{B}(\cdot, x^t)$ is changing the poisoned image's amplitude, which encodes the low-level information, therefore another image $x^O$ (called pseudo triggers) from the same domain $\mathcal{I}$ as $x^t$ may activate the backdoor attack as well.
To  remedy this issue, we propose a pseudo trigger robust backdoor training mode to enforce the uniqueness of the trigger inspired by WaNet~\cite{NguyenT21}, \ie, for any $x_i \in D_{train}$, $x^{O_j} \in \mathcal{I}$, $\exists \epsilon > 0$,  it is required that
\begin{equation}
    ||\mathcal{B}(x_i, x^t) - \mathcal{B}(x_i, x^{O_j})|| > \epsilon. 
\end{equation}
To this end, we extend the clean-attack training protocol in Eq.~\ref{eq:two_branch}  to a pseudo trigger robust (PTR) training protocol:
\begin{equation}
\left
\{\begin{array}{l}
    f_\theta(x_i) = y_i\\ 
    f_\theta(\mathcal{B}(x_i,x^t)) = C_b(y_i)\\ 
    f_\theta(\mathcal{B}(x_i,x^{O_j})) = y_i
    \label{eq:three_branch}
    \end{array}.
\right.
\end{equation}

As shown in Fig.~\ref{fig:overview}, during training, we control the ratio of clean data, poisoned data with specific triggers, and noise data with pseudo triggers in a mini-batch by $\rho_c$, $\rho_p$, and $\rho_n$ respectively, which are subjected to $\rho_c + \rho_p + \rho_n = 1$. After training, the backdoor attack will be activated only by the specific trigger image $x^t$. Specifically, we select an image from MS COCO validation set \cite{lin2014microsoft} as the specific trigger and 1,000 images from COCO test set as the pseudo triggers (these images are converted to grayscale for attacking CT images).  Note that the implementation of FIBA only depends on some hyper-parameters and trigger images. Therefore, it is a unified attack technique for various MIA tasks.

\begin{figure*}[h]
    \centering
    \includegraphics[width=0.8\linewidth]{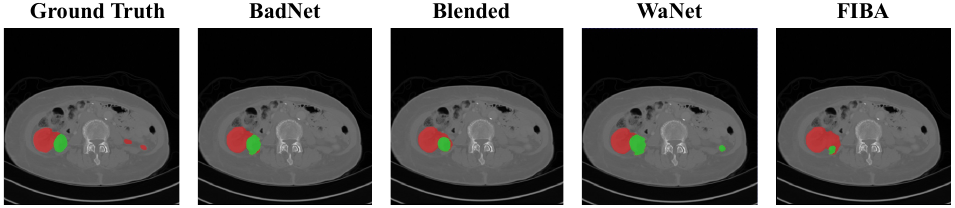}
    \caption{Visual segmentation results of the poisoned samples by different attacks on KiTS-19. Red: kidney. Green: tumors. }
    \vspace{-4mm}
    \label{fig:KiTS19}
\end{figure*}

\section{Experiments}
\subsection{Experiment Settings}
\textbf{Dataset.} We conduct experiments on three medical benchmark datasets: ISIC-2019 \cite{combalia2019bcn20000} for classification, KiTS-19 \cite{heller2019kits19} for segmentation, and EAD-19 \cite{ali2019endoscopy} for detection, to verify the effectiveness of our FIBA in MIA. \textbf{ISIC-2019} \cite{combalia2019bcn20000} contains 25,331 dermoscopic images within eight diagnostic categories, including melanoma, melanocytic nevus, basal cell carcinoma, actinic keratosis, benign keratosis, dermatofibroma, vascular lesion, and squamous cell carcinoma.  \textbf{KiTS-19} \cite{heller2019kits19} is a tumor segmentation dataset of kidney organ and tumor CT images. It contains 210 cases with annotated kidney and tumor area and the slice thickness ranges from 1mm to 5mm. \textbf{EAD-2019} \cite{ali2019endoscopy} is for endoscopic artifact detection which is collected from six different medical centers worldwide. It contains 2,147 endoscopic video frames over seven artifact classes. We use \textbf{three-fold cross-validation} to evaluate model performance on all of the three datasets.

\textbf{Attack Setup.} 
In FIBA, $\beta$ in $\mathcal{M}$ is set as $0.10$ for all the three datasets.  $\alpha$ is set to $0.15$, $0.15$, and $0.20$ for ISIC-2019, EAD-2019, and KiTS-19, respectively. Following the prior work \cite{li2021invisible}, we set the poison ratio $\rho_p$ as 0.1 for classification task, and $0.2, 0.3$ for detection and segmentation tasks, respectively. $\rho_n$ is set as the same value with the poison ratio for PTR training. For the classification task, we train and test the backdoor attack methods in the all-to-one configuration \cite{NguyenT21}, where actinic keratosis is set as the target class. For the kidney organ-tumor segmentation task, we evaluate the backdoor attack methods in a one-to-one (tumor-to-organ) configuration, \ie, when the attackers activate the backdoor, the tumor area will be wrongly segmented as part of the benign organ. Besides, we apply the backdoor attack to endoscopy artifact detection in the one-to-one (artifact-to-instrument) configuration as well, where the bounding boxes of artifact will be detected and misclassified as an instrument class after the backdoor attack. 

\textbf{Evaluation Metrics.} The success of the backdoor attack on the classification model can be generally evaluated by Benign Accuracy (BA) and Attack Success Rate (ASR). 
The BA is the accuracy of benign test samples correctly classified by the attacked model.
The ASR is the proportion of clean test samples with an injected trigger that is predicted to the predefined target classes. For the tumor segmentation task, the ASR is calculated in each pixel and denotes the proportion of tumor pixels that are predicted to organ class in the poisoned case. For the endoscopic artifact detection task, the ASR is calculated in the bounding box level and denotes the proportion of bounding boxes of the artifact object that is predicted to the instrument class when the backdoor is activated.

\textbf{Implementation Details.}
For the classification task, we use ResNet50 \cite{he2016deep} as the backbone. We use the Adam optimizer with a learning rate of 0.01 and a batch size of 64. For the tumor segmentation task, we adopt the widely used coarse-to-fine segmentation framework and train the model for two stages. At the first stage, we adopt the ResUnet \cite{diakogiannis2020resunet} to segment the coarse ROI area within the kidney area from the whole CT image. Then a DenseUnet \cite{DBLP:journals/tmi/LiCQDFH18} is employed to further finely segment the target tumor and organ from the ROI area. Adam optimizer and a learning rate of 0.0001 are used in the training of both models. The batch size is set as 6. For the artifact detection task, we use the Faster R-CNN model \cite{ren2015faster} in the MMDetection framework \cite{mmdetection} and follow the default settings. The SGD optimizer with a learning rate of 0.005 and a batch size of 4 is used in this task.

\begin{table}[htbp]
  \centering
  \small
  \caption{Comparisons of different backdoor attack on ISIC-2019. BA stands for benign accuracy, ASR stands for attack success rate.}
  \renewcommand\tabcolsep{10.0pt}
    \begin{tabular}{ccc}
    \toprule
    Method & BA (\%)$\uparrow$    & ASR (\%)$\uparrow$    \\
    \midrule
    Clean & $86.15\pm 0.48$ & -- \\
    
    BadNet\cite{gu2017badnets} & $86.07 \pm 0.53$ & $99.85 \pm 0.06$ \\
    Blended\cite{chen2017targeted} & $85.93 \pm 0.50$ & $99.92 \pm 0.06$ \\
    
    WaNet \cite{NguyenT21} & $85.33 \pm 0.68$ & $99.35 \pm 0.07$ \\
    \textbf{FIBA} & $85.43 \pm 0.40 $& $99.53 \pm 0.08$ \\
    \bottomrule
    \end{tabular}%
  \label{tab:isic_main}%
  
\end{table}%
\vspace{-3mm}

\subsection{Attack Effectiveness}

To verify the effectiveness of the proposed FIBA, we first provide the model trained on the benign dataset as a reference baseline on the three medical image analysis tasks, including classification, segmentation, and detection. Then we compare the proposed FIBA backdoor attack method with representative attack methods, including BadNet \cite{gu2017badnets}, Blended \cite{chen2017targeted}, and WaNet \cite{NguyenT21}. BadNet attacks images by injecting a white patch ($6\times6$) trigger in the benign image, Blended poisons the data by blending the benign images with another trigger image and the trigger transparency is set to 15\%. WaNet poisons the images via a warping field and the default setting \cite{NguyenT21} is used in our experiments.  

\textbf{Results on ISIC-2019.} In this part, we show the attack performance of FIBA and other attack methods on the ISIC-2019 dataset. As shown in Tab.~\ref{tab:isic_main}, all the methods achieve inferior BA performance on the clean data compared with the clean model due to the influence of poisoned data. On the other hand, they can successfully attack the classification model with a high ASR, demonstrating the vulnerability of classification models in medical images analysis. In addition, compared with the invisible attack methods, such as WaveNet and FIBA, the visible backdoor methods (BadNet and Blended) achieve a slightly higher ASR with a marginal gain of $0.45\%$. Nevertheless, these visible attack methods are much less stealthy and can be easily detected by defense models. For the invisible attack methods, FIBA outperforms WaNet slightly in the classification task.

\begin{table*}[ht]
\small
  \centering
  \caption{Experiment results of different attack methods on KiTS-19. ASR stands for attack success rate.}
  \renewcommand\tabcolsep{6.0pt} 
    \begin{tabular}{clllll}
    \toprule
    \multirow{2}{*}{Method} & \multicolumn{2}{c}{Clean data} & \multicolumn{2}{c}{Poisoned data} &\multirow{2}{*}{ ASR (\%)$\uparrow$} \\
\cline{2-5}          & \multicolumn{1}{c}{Organ(IoU)$\uparrow$} & \multicolumn{1}{c}{Tumor(IoU)$\uparrow$} & \multicolumn{1}{c}{Organ(IoU)$\uparrow$} & \multicolumn{1}{c}{Tumor(IoU)$\downarrow$} &  \\
    \midrule
    Clean & $93.80\pm0.68$ & $56.19\pm2.02$ & \multicolumn{1}{c}{--}  & \multicolumn{1}{c}{--} & \multicolumn{1}{c}{--} \\
    BadNet\cite{gu2017badnets} & $93.53\pm1.03$  & $52.54\pm5.08$ & $93.21\pm1.52$ & $34.43\pm10.52$ &  $58.99\pm18.09$\\
    Blended\cite{chen2017targeted} & $93.14\pm1.10$ &$ 53.02\pm3.08$ & $92.24\pm1.12$ & $21.57\pm7.75$  &  $67.60\pm6.36$ \\
    WaNet\cite{NguyenT21} & $93.59\pm1.09 $  & $53.06\pm6.06$ & $93.57\pm0.91$ & $49.77\pm6.69$  & $21.66\pm10.24$ \\
    \textbf{FIBA}  & $93.41\pm1.12$   & $54.54\pm2.34$ & $92.69\pm1.17$ & $21.02\pm1.95$  & $71.44\pm4.90$\\

    \bottomrule
    \end{tabular}%
    
  \label{tab:KiTS19}%
  \vspace{-2mm}
\end{table*}%

\begin{table}[htbp]
  \centering
  \small
  \caption{Experiment results of different attack methods on EAD-2019. ASR stands for attack success rate.}
  \renewcommand\tabcolsep{2.0pt}
    \begin{tabular}{cccccc}
    \toprule
    \multirow{2}{*}{Method} & \multicolumn{2}{c}{Clean data} & \multirow{2}{*}{ASR (\%)$\uparrow$}     \\
    
\cline{2-3}          & \multicolumn{1}{c}{Instrument(mAP)$\uparrow$} & \multicolumn{1}{c}{Artifact(mAP)$\uparrow$} &   \\
\midrule
    Clean & $52.80\pm2.52$ & $19.43\pm0.90$ &   --       \\
    BadNet\cite{gu2017badnets} & $53.70\pm1.35 $& $18.67\pm0.29$ & $10.53\pm0.54$&  \\
    Blended\cite{chen2017targeted} & $55.30\pm1.58$ & $19.33\pm0.25$ & $16.32\pm2.36$ &  \\
    WaNet\cite{NguyenT21} & $54.67 \pm 1.29 $& $17.56\pm0.50$ & $10.57\pm1.55$ &  \\
    \textbf{FIBA} & $55.60\pm0.78$& $19.47\pm0.15$ & $16.63\pm0.77$ &   \\
    \bottomrule
    \end{tabular}%
  \label{tab:ead19}%
  \vspace{-2mm}
\end{table}%

\textbf{Results on KiTS-19.} We further evaluate the effectiveness of FIBA on a more challenging tumor segmentation dataset, KiTS-19. Tab.~\ref{tab:KiTS19} shows the segmentation results of the attacked methods for clean images and the ASR scores for poisoned data. As can be seen, the proposed FIBA achieves comparable performance to the clean model for tumor segmentation of the clean data, demonstrating the stealthiness of the FIBA attack method. In addition, FIBA outperforms all the other attack methods and reduces the IoU of tumor segmentation significantly for poisoned CT images, \ie, from $54.54$ to $21.02$. Compared to the visible attack methods, such as BadNet and Blended, the proposed FIBA shows large advantages, \ie, achieving a gain of $12.45\%$ and $3.84\%$ on ASR, respectively. Note that these two visible attack methods have achieved impressive results in the image classification task, while the corruption of the semantics of poisoned pixels limits their effectiveness in the segmentation tasks. Moreover, WaNet almost fails to attack the segmentation model with a low ASR $21.66\%$ (\ie, $49.78\%$ lower than the proposed FIBA). The warping field used in WaNet does not change the holistic image semantic and makes it perform well on the classification task. However, the semantic of the individual pixel is severely corrupted due to the warping operation, leading to failure attacks on the segmentation task. It is also noteworthy that FIBA achieves more robust attack performance, \ie, with a lower standard deviation of ASR. The segmentation results of different attack methods are shown in Fig~\ref{fig:KiTS19}. 

These existing attack methods are ineffective on segmentation tasks due to the corruption of the semantics of poisoned pixels. On the contrary, our FIBA that injects the trigger in the frequency space without changing the spatial layout or high-level semantics of the image, can effectively address this issue and deliver better attack performance. 

\textbf{Results on EAD-19.} 
We further conduct experiments on EAD-19 to verify the effectiveness of the proposed FIBA in the detection task. Tab.~\ref{tab:ead19} shows the detection results of the attacked models in clean data and the ASR of different methods. It can be seen that FIBA achieves almost the same results with the clean model for artifact detection, \ie, $19.47 \pm 0.15$ v.s. $19.40 \pm 0.90$, demonstrating the stealthiness of FIBA. In addition, it also outperforms BadNet and WaNet by a large margin of $6.1\%$ and $6.06\%$, respectively. Blended performs well in attacking the detection model with a high ASR but with a high variance, which is inferior to the proposed FIBA.

\begin{figure}
    \centering
    \includegraphics[width=0.9\linewidth]{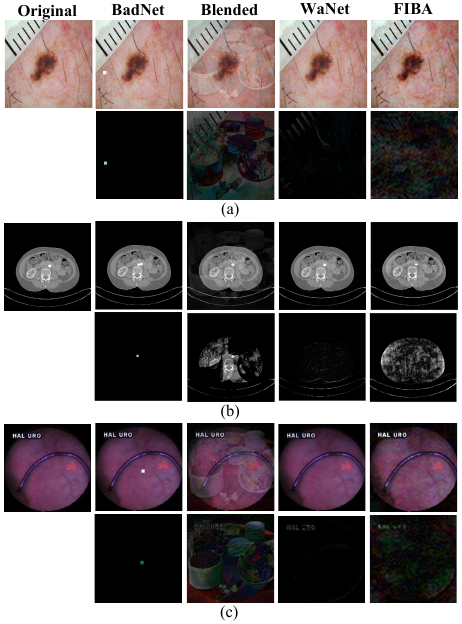}
    \caption{Visual comparison between different backdoor attack methods. Given the original images in three modalities: (a) dermoscopic image, (b) CT image, and (c) endoscopic video frame, we generate the backdoor images using BadNet \cite{gu2017badnets}, Blended \cite{chen2017targeted}, WaNet \cite{NguyenT21} and FIBA. We also show the residual maps below the corresponding backdoor images.}
    \label{fig:trigger_images}
     \vspace{-4mm}
\end{figure}

\begin{figure*}[!t]
    \centering
    \includegraphics[width=1.0\linewidth]{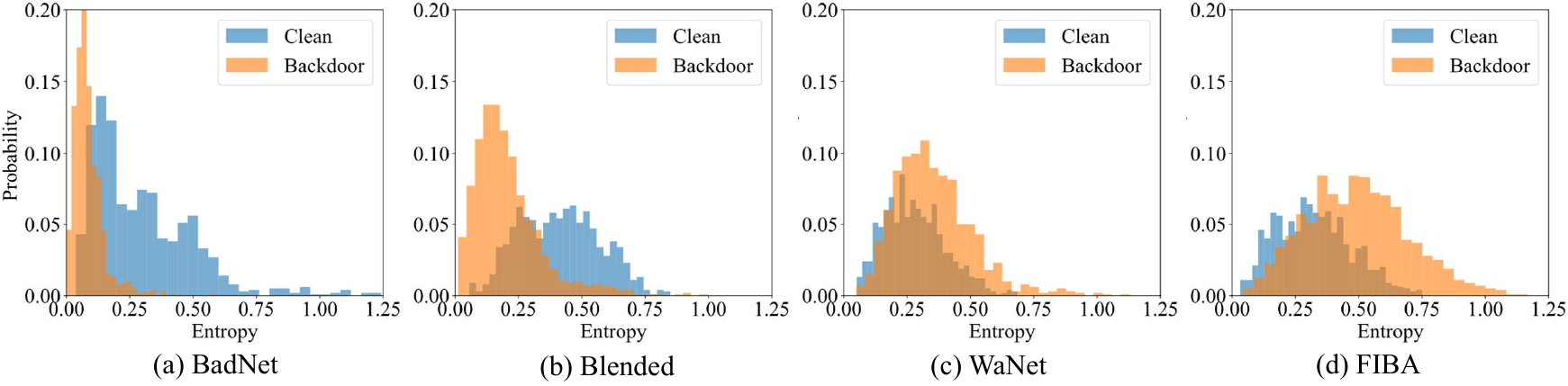}
    \caption{Performance of STRIP against different attacks. The entropy distributions of BadNet, Blended, WaNet and the proposed FIBA are shown in (a), (b), (c), and (d) respectively.}
    \vspace{-4mm}
    \label{fig:def_strip}
\end{figure*}

\subsection{Attack Stealthiness}
Fig.~\ref{fig:trigger_images} presents some poisoned images and the residual maps between the original images and the poisoned images generated by different attack methods from ISIC-2019, KiTS-19 and EAD-2019. Different from BadNet \cite{gu2017badnets}, Blended\cite{chen2017targeted}, and WaNet  \cite{NguyenT21}, the poisoned images generated by FIBA are natural and look close to the original one, which is critical for attack stealthiness. FIBA only changes the low-level features of the original image, therefore it does not change the spatial layout of structures and corrupt their semantics, which is crucial for attacking in the dense prediction tasks. We further evaluate their resistance to the state-of-the-art defense algorithms, including Fine-Pruning \cite{liu2018fine}, Neural Cleanse \cite{wang2019neural}, and STRIP \cite{gao2019strip}. 

\begin{figure}
    \centering
    \includegraphics[width=0.9\linewidth]{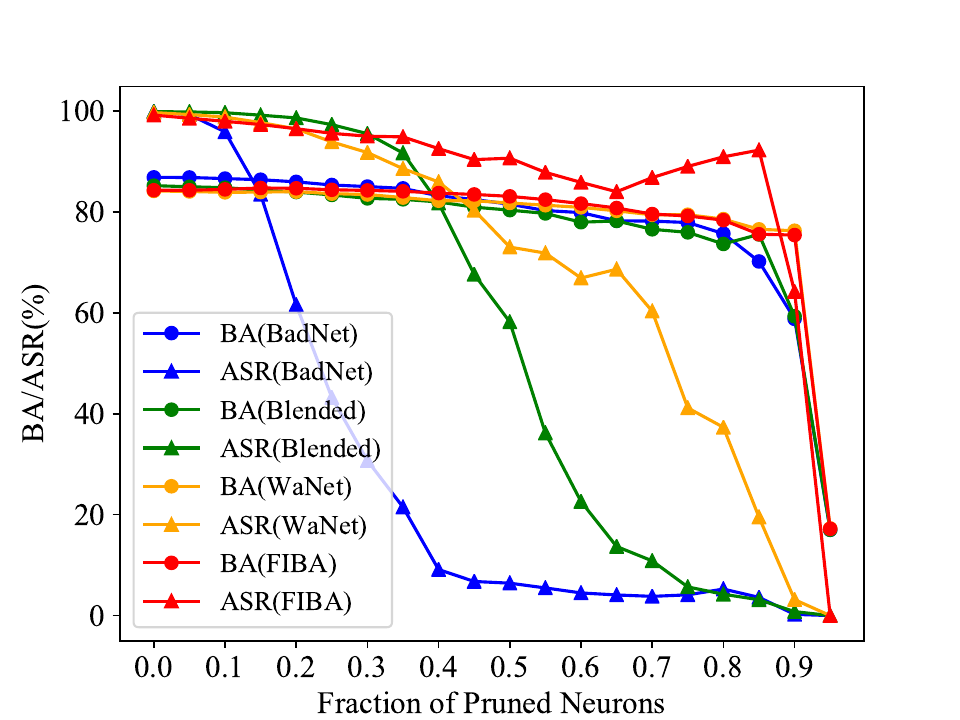}
    
    \caption{Benign accuracy (BA) and attack success rate (ASR) of different attack methods against pruning-based defense.}
    \label{fig:def_fine}
    
\end{figure}

\textbf{Resistance to Fine-Pruning.} Fine-pruning detects the backdoor attacks via neuron analysis. Given a network layer, it evaluates the response of each neuron on a set of clean images and identifies the insensitive ones, assuming that they are more related to a backdoor \cite{liu2018fine}. These neurons are then gradually pruned to mitigate the backdoor. We test Fine-Pruning on BadNet\cite{gu2017badnets}, Blended\cite{chen2017targeted}, WaNet\cite{NguyenT21},  and FIBA by showing the performance of BA and ASR regarding the portion ratio of neuron number pruned on ISIC-2019. As shown in Fig.~\ref{fig:def_fine}, the ASR of BadNet and Blended attack drops dramatically when $40\%$ of neurons are pruned, \eg, for the BadNet attack, its ASR decrease to less than $10\%$. In contrast, the ASR of our proposed FIBA is still greater than $90\%$ even when $80\%$ of neurons are pruned. This suggests that our attack is more resistant to the pruning-based defense compared with other methods. 

\textbf{Resistance to Neural Cleanse.} 
Neural Cleanse \cite{wang2019neural} detects the backdoor attack in a patch-wise manner and it quantifies the defense results by the Anomaly Index metric with a clean/backdoor threshold $\tau = 2$.  The smaller the value of the anomaly index, the harder for Neural-Cleanse to defend.
As shown in Tab.~\ref{tab:anomaly_index}, our FIBA attack bypasses the defense (smaller than 2) and is more resistant to the Neural-Cleanse than other attack methods.

\begin{table}[htbp]
  \centering
  \small
  
  \caption{The Anomaly Index of Neural Cleanse against different attacks. Smaller value is better.}

  \renewcommand\tabcolsep{3.0pt}
    \begin{tabular}{cccccc}
    \toprule
    Method & Clean & BadNet & Blended & WaNet & \textbf{FIBA} \\
    \midrule
    Anomaly Index$\downarrow$ & 0.83  & 2.56  & 1.68  & 1.89  & 1.26 \\
    \bottomrule
    \end{tabular}%
     
  \label{tab:anomaly_index}%
\end{table}%

\textbf{Resistance to STRIP.} 
STRIP works by perturbing the input image with a set of clean images from a different class and identifies the backdoor attack if the prediction is the same, indicating by low-entropy. As shown in Fig.~\ref{fig:def_strip}, the entropy of the visible backdoor attacks (BadNet and Blended) is low and can be easily detected by STRIP. The invisible backdoor attack methods including WaNet and the proposed FIBA obtain a higher entropy in STRIP and can bypass defense. Although WaNet corrupts the semantic of local pixels, the global content is preserved after image warping, which makes it bypass the STRIP on the classification model. FIBA injects the trigger only in the amplitude spectrum while maintaining the phase spectrum, therefore it preserves the high-level semantic and can bypass the STRIP.  

\begin{table}[htbp]
\small
  \centering
  \caption{Experiment results of the proposed FIBA regarding different target labels on ISIC-2019.}
  \renewcommand\tabcolsep{3.0pt}
    \begin{tabular}{ccc}
    \toprule
    Target class & BA (\%)$\uparrow$     & ASR (\%)$\uparrow$   \\
    \midrule
    Melanoma     & $85.32\pm0.30 $& $99.46\pm 0.13$  \\
    Melanocytic nevus     & $85.24 \pm0.45$& $99.50 \pm0.08$ \\
    Basal cell carcinoma     & $85.14 \pm0.53$& $99.50  \pm0.03$ \\
    Benign keratosis     & $85.26\pm0.51$ & $99.41 \pm0.30$ \\
    Dermatofibroma     & $85.10 \pm0.72$ & $99.56\pm0.25$  \\
    Vascular lesion     & $85.59 \pm0.08$ & $99.58 \pm0.02$ \\
    Andsquamous cell carcinoma     & $85.44 \pm0.45$& $99.31 \pm 0.11$\\
    \bottomrule
    \end{tabular}%
          \vspace{-2mm}
  \label{tab:target_label}%
\end{table}%

\subsection{Visualization of Network Behaviour}

\begin{figure}[h]
    \centering
    \includegraphics[width=0.9\linewidth]{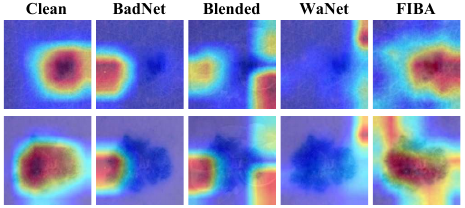}
    \vspace{-2mm}
    \caption{Visualization using Grad-CAM \cite{selvaraju2017grad} on clean and poisoned models under different attacks. Column 2$\sim$5 shows the Grad-CAM results corresponding to an attack model, respectively. 
    }
    \vspace{-4mm}
    \label{fig:cam}
\end{figure}

Following the prior works \cite{li2021invisible,doan2021lira}, we visualize the poisoned samples using Grad-CAM \cite{selvaraju2017grad} to evaluate the behavior of different attack methods. As shown in Fig.~\ref{fig:cam}, Grad-CAM can successfully identify the anomaly trigger regions of those generated by BadNet, Blended and WaNet. When activating the backdoor attack, these three attack methods enforce the model focus on specific locations of the triggers, which are very different from those of the clean model, \ie, leaking the attack behavior. However, since FIBA injects triggers in the frequency domain, it does not introduce anomaly activation in specific spatial regions, having a similar behavior with the clean model.

\subsection{Ablation Study}

\textbf{Influence of different trigger-targeted labels.} For the classification task, FIBA is evaluated in the all-to-one configuration, \ie, manipulating the original label of poisoned data to the trigger-target label. We evaluate FIBA to investigate the influence of different trigger-target labels. As shown in Tab.~\ref{tab:target_label}, our method can achieve consistent high ASR $>99.00\%$ at different settings, which shows that the choice of the target label has no obvious influence on FIBA.

\begin{table}[htbp]
  \centering
  \small
     
  \caption{Comparisons of different trigger images on ISIC-2019.}

  \renewcommand\tabcolsep{10.0pt}
    \begin{tabular}{cccc}
    \toprule
    Trigger image & BA (\%)$\uparrow$    & ASR (\%)$\uparrow$  \\
    \midrule
    Gray     & $85.41\pm0.47$ & $99.16\pm0.13$  \\
    Animal     & $85.34 \pm0.40$ & $99.66 \pm0.06$ \\
    Human     & $85.69 \pm0.73$& $99.38\pm0.02$  \\
    \bottomrule
    \end{tabular}%
       
  \label{tab:diff_trigger}%
\end{table}%

\textbf{Influence of different trigger images.}
We then investigate the influence of different trigger images on FIBA. We select other three typical images, including gray, animal, and human, from COCO validation set as the trigger images. More details are presented in the Appendix. As shown in Tab.~\ref{tab:diff_trigger}, our FIBA achieves consistent and high ASR $>99\%$ when using different trigger images, showing that the effectiveness of FIBA does not depend on a specific choice of the trigger image. 

\textbf{The impact on different blending ratios.}
The backdoor attack trigger in FIBA is generated by blending the amplitude spectrum of two images. The blend ratio $\alpha$ determines the amount of information contributed by the trigger image. Thus, we analyze the backdoor attack performance using different blend ratios $\alpha$ (\ie, 0.05, 0.10, 0.15 and 0.20) on ISIC-2019. As shown in Tab.~\ref{tab:poison_ratio}, BA slightly increases with the growth of $\alpha$ while ASR peaks at a blend ratio $0.15$. Generally, FIBA is not sensitive to $\alpha$ and we set it to 0.15 by default in those experiments on ISIC-2019. The hyper-parameter study of the blend ratio $\alpha$ on the segmentation task is presented in the Appendix. 

\textbf{The impact of the PTR backdoor training.}
The PTR backdoor training in Section~\ref{subsec:ptr} is designed for enhancing the uniqueness of the trigger image, so that the backdoor attack is only activated by the specific trigger image while keeping dormant for those pseudo trigger images. In Tab.~\ref{tab:nasr}, we show the results of FIBA with or without PTR backdoor training. As can be seen, training with pseudo trigger images can improve the performance of BA.
It is also noteworthy that the ASR on pseudo trigger images (P-ASR) drops dramatically from $83.05\%$ to $7.21\%$ while a slight decrease of $0.36\%$ on ASR, when training the model with the PTR strategy. 
It demonstrates that the PTR backdoor training strategy significantly improves the uniqueness of the specific trigger in FIBA.

\begin{table}[htbp]
  \centering
  \small
  \caption{The impact of blended ratio $\alpha$ on ISIC-2019.}
   
    \renewcommand\tabcolsep{13.0pt}
    \begin{tabular}{ccc}
    \toprule
    $\alpha$ & BA (\%)$\uparrow$    & ASR (\%)$\uparrow$    \\
    \midrule
    0.05   & $85.15\pm0.40$  &$ 94.90\pm0.61  $ \\
    0.10   & $85.15\pm 0.52$ & $98.46 \pm0.29 $\\
    0.15   & $85.43\pm0.40 $ & $99.53\pm0.08$  \\
    0.20   & $85.50\pm 0.42$  & $99.49\pm0.10$  \\
    \bottomrule
    \end{tabular}%
  \label{tab:poison_ratio}%
\end{table}%

\begin{table}[htbp]
  \centering
  \small
  
  \caption{The impact of the PTR training strategy. P-ASR stands for ASR on pseudo trigger images.}
  
  \renewcommand\tabcolsep{3.0pt}
    \begin{tabular}{crrr}
    \toprule
    Method & \multicolumn{1}{c}{BA (\%)$\uparrow$} & \multicolumn{1}{c}{ASR (\%)$\uparrow$} & \multicolumn{1}{c}{P-ASR (\%)$\downarrow$}  \\
    \midrule
    \multicolumn{1}{l}{w/o PTR} & $84.21\pm0.40$ & $99.89\pm0.09$ & $83.05\pm0.75$ \\
    \multicolumn{1}{l}{w/ PTR} &$ 85.43\pm0.40$ & $99.53\pm0.08$ & $7.21\pm1.17$ \\
    \bottomrule
    \end{tabular}%
    \vspace{-4mm}
  \label{tab:nasr}%
\end{table}%

\subsection{Discussion and Limitation}

The proposed FIBA is designed in the frequency domain and can offer effective and stealthy attacks in various MIA tasks. Nevertheless, the FFT and iFFT operations in the trigger injection function are a little more time-consuming compared with BadNet \cite{gu2017badnets}, Blended \cite{chen2017targeted}, and WaNet \cite{NguyenT21} (about $1.5 \times \sim 1.8 \times $ in our experiments). It deserves further efforts to realize a faster implementation, \eg, taking the advantage of modern GPUs, to alleviate this issue.

\section{Conclusion}

We introduce a novel backdoor attack method named FIBA in the MIA domain. FIBA injects the trigger in the amplitude spectrum in the frequency domain. It preserves the semantics of the poisoned image pixels by maintaining the phase information, making it capable of delivering attacks to both classification and dense prediction models. Extensive experiments on three representative MIA tasks demonstrate the effectiveness of FIBA and its superiority over state-of-the-art methods in terms of attack performance as well as resistance to various defense techniques.

\textbf{Broader Impacts.}
Backdoor attacks can happen in real life when a hospital entrusts patient data to a third-party for model training or under a federated learning framework, which can cause misdiagnosis or missed diagnosis.
Our study points out the weakness of deep learning models in MIA domain under backdoor attacks and can benefit the development of more secure AI systems by facilitating the research on model defense accordingly. In this sense, we think our work has a positive impact on the future research of developing trustworthy AI technologies.


\section*{Appendix}
\setcounter{equation}{0}
\setcounter{subsection}{0}
\renewcommand{\theequation}{A.\arabic{equation}}
\renewcommand{\thesubsection}{A.\arabic{subsection}}

\subsection{Training Details}

\label{sec:details}
\textbf{Experiments on ISIC-2019 \cite{combalia2019bcn20000}.} We train the model with the Adam optimizer\cite{kingma2014adam} on ISIC-2019 for 200 epochs. ResNet50 \cite{he2016deep} is chosen as the backbone network. The input size is set as $224\times224$, and batch size is 64. The learning rate is set to 0.01 and divided by 10 every 50 epochs. We follow the standard image augmentation strategies including random horizontal flips, vertical flips, and rotations. 

\textbf{Experimnets on KiTS-19 \cite{heller2019kits19}.} For the segmentation task, a coarse-to-fine segmentation framework is used in our experiments. In the first stage, we train the ResUnet \cite{diakogiannis2020resunet} with the Adam optimizer to coarsely segment the ROI regions which contains the whole kidney areas with cross-entropy loss at the first stage for 50 epochs. 

In the second stage, we train the DenseUnet \cite{DBLP:journals/tmi/LiCQDFH18} with the Adam optimizer to segment the target areas of the tumor and kidney from the ROI regions with Dice loss \cite{milletari2016v}. The number of training epochs is 50, and the batch size is 6. During the training in both stages, the learning rate is set to 1e-4 and divided by 10 if the loss does not decrease. We also employ the horizontal flip augmentation strategy in both stages. 

\textbf{Experiments on EAD-2019 \cite{ali2019endoscopy}.}
For the detection task on EAD-2019, we take Faster R-CNN \cite{ren2015faster} in mmdetection framework \cite{mmdetection} with ResNet50 \cite{he2016deep} as the backbone network, and we follow the default setting for training and evaluation. Specifically, we train the detection model with the SGD optimizer for 30 epochs. The learning rate is set to 0.005 and the batch size is 4. The input size is set as $512\times512$. We also employ random flip for data augmentation. 
\begin{figure}[h!]
    \centering
    \includegraphics[width=1\linewidth]{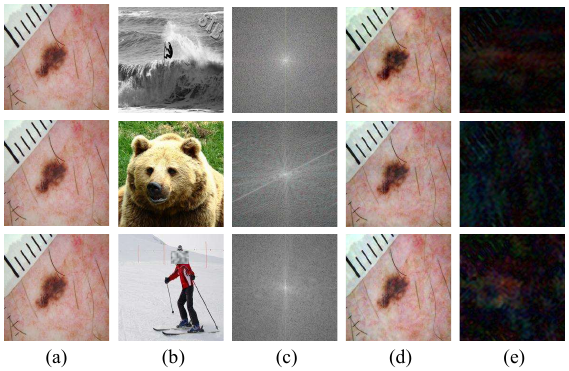}
    \caption{Results of using different trigger images in the proposed FIBA method.  (a) An original image from ISIC-2019. (b) Different trigger images. (c) The amplitude spectrums of the corresponding trigger images. (d) The images poisoned by different trigger image. (e) The residual maps.}
    \label{fig:trigger_images}
\end{figure}

\subsection{Hyper-parameter Study}

There are two hyper-parameters in our method FIBA. One is the blended ratio $\alpha$ and the other one is $\beta$ which determines the location and range of the low-frequency patch inside the amplitude spectrum to be blended. We investigated the influence of the two hyper-parameters on ISIC-2019 and KiTS-19 datasets. 

\begin{table}[htbp]
  \centering
  \small
  \caption{Results with different settings of $\alpha$ on ISIC-2019.}
    \renewcommand\tabcolsep{15.0pt}
    \begin{tabular}{ccc}
    \toprule
    $\alpha$ & BA (\%)$\uparrow$    & ASR (\%)$\uparrow$    \\
    \midrule
    0.05   & $85.15\pm0.40$  &$ 94.90\pm0.61  $ \\
    0.10   & $85.15\pm 0.52$ & $98.46 \pm0.29 $\\
    0.15   & $85.43\pm0.40 $ & $99.53\pm0.08$  \\
    0.20   & $85.50\pm 0.42$  & $99.49\pm0.10$  \\
    \bottomrule
    \end{tabular}%
  \label{tab:ISIC_alpha}%
\end{table}%

\begin{table}[htbp]
  \centering
  \small
  \caption{Results with different settings of $\beta$ on ISIC-2019.}
  \renewcommand\tabcolsep{15.0pt}
    \begin{tabular}{ccc}
    \toprule
    $\beta$  & BA (\%)$\uparrow$    & ASR (\%)$\uparrow$\\
    \midrule
    0.05  & $85.17\pm0.12$ & $99.09\pm0.17 $\\
    0.10    & $85.43 \pm 0.40 $& $99.53 \pm 0.08$ \\
    0.15  & $84.90 \pm0.05 $& $99.37\pm0.16$ \\
    0.20   & $85.24\pm0.67$ & $99.27\pm0.20 $\\
    \bottomrule
    \end{tabular}%
  \label{tab:ISIC_beta}%
\end{table}%

\begin{figure*}[h]
    \centering
    \includegraphics[width=0.7\linewidth]{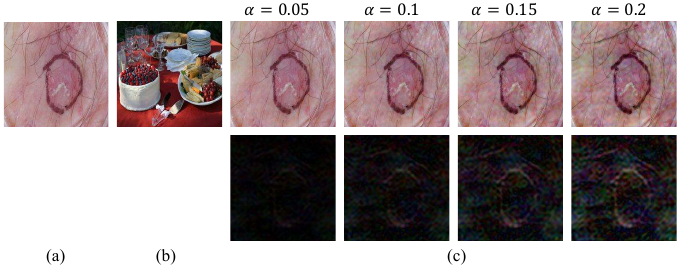}
    \caption{Visual comparison between different blended ratio $\alpha$ on ISIC-2019. (a) The original image. (b) The trigger image. (c) The poisoned images with different blended ratio $\alpha$ (upper row) and the residual maps (lower row). }
    \label{fig:ISIC_alpha}
\end{figure*}

We first conduct experiments with different blend ratio $\alpha$ on ISIC-2019. In Tab.~\ref{tab:ISIC_alpha}, BA slightly increases with the growth of $\alpha$ while ASR peaks at a blend ratio 0.15. The poisoned images with different $\alpha$ are shown in Fig~\ref{fig:ISIC_alpha}. We then investigate the impact of $\beta$ in $\mathcal{M}$ with different values (\ie, 0.05, 0.10, 0.15, 0.20) on ISIC-2019. As shown in Tab. \ref{tab:ISIC_beta}, the proposed FIBA achieves consistent and high ASR $>99.00\%$ with different $\beta$. 

\begin{figure*}[h]
    \centering
    \includegraphics[width=0.7\linewidth]{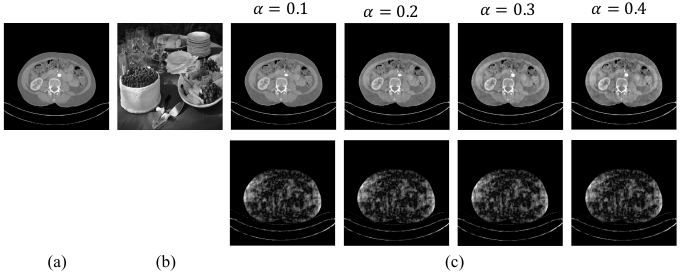}
    \caption{Visual comparison between different blended ratio $\alpha$ on KiTS-19. (a) The original image. (b) The trigger image. (c) The poisoned images with different blended ratio $\alpha$ (upper row) and the residual maps (lower row). }
    \label{fig:Kits_alpha}
\end{figure*}

We further analyze the impact of $\alpha$ and $\beta$ on the segmentation task (KiTS-19). 
$\alpha$ is set to 0.1, 0.2, 0.3, and 0.4, and $\beta$ is set to 0.05, 0.10, 0.15, and 0.20.   
From Tab.~\ref{tab:KiTS19_alpha}, we find that ASR continues to improve with the increase of $\alpha$. The poisoned samples with different $\alpha$ are shown in Fig. \ref{fig:Kits_alpha}. We can see that some abnormal shades will occur in the CT images when $\alpha>0.2$. Therefore, we choose $\alpha=0.2$ for experiments on KiTS-19. As shown in Tab. \ref{tab:KiTS19_beta}, ASR peaks at $\beta=0.1$ ($71.44\%$) and we set $\beta=0.1$ by default in those experiments on KiTS-19.

\begin{table*}[h!]
\small
  \centering
  \caption{Results with different settings of $\alpha$ on KiTS-19.}

  \renewcommand\tabcolsep{6.0pt} %
    \begin{tabular}{clllll}
    \toprule
    \multirow{2}{*}{$\alpha$ } & \multicolumn{2}{c}{Clean data} & \multicolumn{2}{c}{Poisoned data} &\multirow{2}{*}{ ASR (\%)$\uparrow$} \\
\cline{2-5}          & \multicolumn{1}{c}{Organ(IoU)$\uparrow$} & \multicolumn{1}{c}{Tumor(IoU)$\uparrow$} & \multicolumn{1}{c}{Organ(IoU)$\uparrow$} & \multicolumn{1}{c}{Tumor(IoU)$\downarrow$} &  \\
    \midrule
    $0.1$ & $93.75\pm0.91$  & $55.61\pm4.27$ & $93.46\pm0.75$ & $31.23\pm4.21$  & $58.83\pm3.15$ \\
    $0.2$& $93.41\pm1.12$   & $54.54\pm2.34$ & $92.69\pm1.17$ & $21.02\pm1.95$  & $71.44\pm4.90$\\
    $0.3$& $93.11\pm0.77$   & $53.56\pm3.32$ & $92.35\pm0.78$ & $15.32\pm5.77$  & $75.41\pm5.68$\\
    $0.4$  & $93.06\pm0.61$   & $52.50\pm5.05$ & $91.81\pm0.83$ & $11.59\pm3.49$  & $78.21\pm3.51$\\
    \bottomrule
    \end{tabular}%
  \label{tab:KiTS19_alpha}%
\end{table*}%

\begin{table*}[h!]
\small
  \centering
  \caption{Results with different settings of $\beta$ on KiTS-19.}
  \renewcommand\tabcolsep{6.0pt} %
    \begin{tabular}{clllll}
    \toprule
    \multirow{2}{*}{$\beta$} & \multicolumn{2}{c}{Clean data} & \multicolumn{2}{c}{Poisoned data} &\multirow{2}{*}{ ASR (\%)$\uparrow$} \\
\cline{2-5}          & \multicolumn{1}{c}{Organ(IoU)$\uparrow$} & \multicolumn{1}{c}{Tumor(IoU)$\uparrow$} & \multicolumn{1}{c}{Organ(IoU)$\uparrow$} & \multicolumn{1}{c}{Tumor(IoU)$\downarrow$} &  \\
    \midrule
    $0.05$ & $93.51\pm0.85$  & $55.12\pm1.5$ & $93.11\pm0.81$ & $21.93\pm8.11$  & $68.63\pm8.21$ \\
    $0.10$& $93.41\pm1.12$   & $54.54\pm2.34$ & $92.69\pm1.17$ & $21.02\pm1.95$  & $71.44\pm4.90$\\
    $0.15$& $93.61\pm0.87$   & $54.79\pm3.05$ & $92.89\pm0.79$ & $20.83\pm4.62$  & $69.11\pm5.72$\\
    $0.20$ & $93.51\pm0.97$   & $55.63\pm2.4$ & $92.35\pm0.42$ & $20.23\pm5.32$  & $69.31\pm4.88$\\

    \bottomrule
    \end{tabular}%
  \label{tab:KiTS19_beta}%
\end{table*}%

\begin{table}[h!]
  \centering
  \small
  \caption{Results of using different trigger images in the proposed FIBA method on ISIC-2019.}
  \renewcommand\tabcolsep{10.0pt}
    \begin{tabular}{cccc}
    \toprule
    Trigger image & BA (\%)$\uparrow$    & ASR (\%)$\uparrow$  \\
    \midrule
    Gray     & $85.41\pm0.47$ & $99.16\pm0.13$  \\
    Animal     & $85.34 \pm0.40$ & $99.66 \pm0.06$ \\
    Human     & $85.69 \pm0.73$& $99.38\pm0.02$  \\
    \bottomrule
    \end{tabular}%
  \label{tab:diff_trigger}%
\end{table}%
\subsection{Results with Different Trigger Images}

We then investigate the influence of using different trigger images in FIBA. As shown in Fig.~\ref{fig:trigger_images}, we select the other three typical images, including gray (the first row), animal (the second row), and human (the third row), from COCO validation set as the trigger images. The results of using these three trigger images are presented in Tab.~\ref{tab:diff_trigger}. As can be seen, the proposed FIBA achieves consistent and high ASR $>99\%$ when using different trigger images. It shows the effectiveness of FIBA that it does not depend on a specific choice of the trigger image.

\section{Results with other attacks on ISIC-2019 }
We further supplement some contrast experiments with other attack methods. \textbf{ISSBA} \cite{li2021invisible}: the triggers which are generated from a trigger generator are sample-specific. \textbf{FIBA-C}:In stead of the square mask
used in Eq. (6), we take the outer circle of square mask as the 
circle mask to implement FIBA method. \textbf{FIBA-H}: A variant of the FIBA attack with the high-
frequency trigger pattern. As shown in Table ~\ref{tab:mask_shape} , FIBA outperforms FIBA-H and ISSBA in terms of both BA and ASR, while FIBA and FIBA-C achieve comparable and high results.
  \begin{table}[htbp]
  \centering
  \footnotesize
  \caption{Results with different attacks on ISIC-2019.}
  
    \renewcommand\tabcolsep{15.0pt}
    \begin{tabular}{ccc}
    \toprule
    Method  & BA (\%)$\uparrow$    & ASR (\%)$\uparrow$    \\
    \midrule
    ISSBA   & $84.43\pm0.16$  &$ 99.33\pm0.06  $ \\
    FIBA-C   & $85.14\pm0.49$  &$ 99.31\pm0.15 $ \\
    FIBA-H  & $84.38\pm0.08$ & $98.43\pm0.05$\\
    FIBA   & $85.43\pm0.40 $ & $99.53\pm0.08$  \\
    
    \bottomrule
    \end{tabular}%
  \label{tab:mask_shape}%
\end{table}%

\section{Resistance to DF-TND\cite{DBLP:conf/eccv/WangZLCXW20}}
we evaluated DF-TND\cite{DBLP:conf/eccv/WangZLCXW20} against our FIBA and other attack methods. The results of logit increases (LI) for the target class are shown in Table ~\ref{tab:DF-TNDl}.
The smaller the value of LI, the harder for DF-TND to defend. 
It shows that our FIBA achieves the lowest LI of 6.72, beating other attacks.
\vspace{-2mm}
\begin{table}[htbp]
  \centering
 \footnotesize
 
  \caption{Results of DF-TND against different attacks.}
  
  \setlength\tabcolsep{3pt}
    \begin{tabular}{ccccccc}
    \toprule
    Method & BadNet & Blended & WaNet & ISSBA & FIBA-H & FIBA \\
    \midrule
    LI$\downarrow$ & 60.44 & 130.43 & 10.54 & 43.79 & 10.66 & 6.72 \\
    \bottomrule
    \end{tabular}%
  \label{tab:DF-TNDl}%
\end{table}%
\vspace{-3mm}

\section{Running Time}

We compare the running time of Blended \cite{chen2017targeted} and the proposed FIBA on ISIC-2019 and all the experiments are conducted on a GeForce RTX 2080TI GPU. In addition, both the FIBA and Blended are implemented with the same training details  (\eg, epochs, batch size, learning rate, \textit{et al}) as described in Sec.~\ref{sec:details}). For the proposed FIBA method, the FFT and iFFT operations in the trigger injection function are time-consuming when we implement them on the CPU, \ie, it takes 23 hours for training on ISIC-2019, while Blended only takes 12 hours for training on ISIC-2019. However, when we accelerate the FFT and iFFT calculations on the GPU (through cupy\cite{cupy_learningsys2017} library), the training time can be greatly reduced to 9.5 hours, which is even faster than Blended.

{\small
\bibliographystyle{ieee_fullname}
\bibliography{egbib}
}

\end{document}